\documentclass[a4paper,10pt]{article}
\usepackage{spconf,amsmath,graphicx,cite,multirow,multicol,algorithm,algorithmic}
\usepackage{amsfonts}
\usepackage{pgfplots}
\usepackage{relsize}
\usepackage{pgfplotstable}
\pgfplotsset{compat=newest}

\title{L}
\name{P.V. Giampouras, A.A. Rontogiannis,  K.D. Koutroumbas}
\address{IAASARS, National Observatory of Athens, GR-15236, Penteli, Greece}
\begin{document}
%
\title{Low-rank and Sparse NMF for Joint Endmembers' Number Estimation and Blind Unmixing of Hyperspectral Images}
%
%
%
%

\name{Paris V. Giampouras, Athanasios A. Rontogiannis, Konstantinos D. Koutroumbas}
\address{IAASARS, National Observatory of Athens, GR-15236, Penteli, Greece}
%
%

\markboth{Journal of \LaTeX\ Class Files,~Vol.~14, No.~8, August~2015}%
{Shell \MakeLowercase{\textit{et al.}}: Bare Advanced Demo of IEEEtran.cls for IEEE Computer Society Journals}
%



\maketitle
\begin{abstract}
Estimation of the number of endmembers existing in a scene constitutes a critical task in the hyperspectral unmixing process. The accuracy of this estimate plays a crucial role in subsequent  unsupervised unmixing steps i.e., the derivation of the spectral signatures of the endmembers (endmembers' extraction) and the estimation of the abundance fractions of the pixels. A common practice amply followed in literature is to treat endmembers' number estimation and unmixing, independently as two separate tasks, providing the outcome of the former as input to the latter. In this paper, we go beyond this computationally demanding strategy. More precisely, we set forth a multiple constrained optimization framework, which encapsulates endmembers' number estimation and unsupervised unmixing in a single task. This is attained by suitably formulating the problem via a low-rank and sparse nonnegative matrix factorization rationale, where low-rankness is promoted with the use of a sophisticated $\ell_2/\ell_1$ norm penalty term. An alternating proximal algorithm is then proposed for minimizing the emerging cost function. The results obtained by simulated and real data experiments verify the effectiveness of the proposed approach. 
\end{abstract}





%

\section{Introduction}
\label{sec:introduction}
Big imaging data, such as hyperspectral images (HSIs) and video, convey a sheer bulk of information and this makes them valuable tools in a plethora of  applications, with remote sensing being probably  the most prominent one \cite{ma_2014}. One of the HSI processing tasks that has attracted considerable attention lately is that of hyperspectral unmixing (HU). The main goal of HU is to estimate the spectral signatures of the components (called as {\it endmembers}) that form the spectral information of HSIs, along with the fractions of their contribution (called as {\it abundances}) at each pixel. 

A first critical and indispensable step towards performing unmixing is to uncover the true number of endmembers that exist in a given hyperspectral scene. This challenging task (also known as {\it rank estimation} or {\it model order selection}), can be quite daunting in terms of the required computational burden. Several works have come into play for attacking this problem which can be classified into two main categories: a) information theoretic criteria based approaches and b) eigenvalue thresholding methods. As far as the first class is concerned, various approaches have been proposed differing in the criterion used for penalizing an initially overestimated number of  endmembers e.g. Akaike's Information Criterion (AIC), Minimum Description Length (MDL), Bayesian Information Criterion (BIC). On the other hand, methods that belong to the second class include PCA based methods, Neyman-Peyrson detection theory based methods etc., \cite{bioucas2008hyperspectral}. 

The estimate of the number of endmembers by the above-mentioned algorithms is provided -at a second phase- as input to unsupervised unmixing algorithms, whose goal is to extract the endmembers' spectral signatures along with the abundance fractions of the pixels. A vast amount of works have been published in the literature dealing with unsupervised hyperspectral unmixing. The majority of them hinges on the assumption that the underlying mechanism that describes the mixing process is linear. Indeed, the so-termed {\it linear mixing model} (LMM) has been proven to be a reliable approximation, although it neglects non-linear effects met in real situations. In the framework of the LMM, there exist both geometrical,\cite{vca_2005}, and statistical, \cite{Lu_2014}, matrix factorization based approaches for performing unmixing. Among the latter, nonnegative matrix factorization (NMF) based techniques have exhibited a robust behavior offering promising results.

In this paper, we propose a multiple constrained NMF method for {\it simultaneously} a) determining the number of endmembers, b) extracting the endmembers' spectral signatures and c) estimating the abundance values of the pixels. To accomplish this, we introduce a sophisticated low-rank promoting term, which is based on the group-sparsity $\ell_2/\ell_1$ inducing norm. This term penalizes both endmembers' and abundance matrices by enforcing joint sparsity on their columns. This way, we go one step beyond just revealing the rank,  since we further encourage estimation  of the true bases of the column spaces of these matrices. At the same time, sparsity is favored on the abundance matrices, as it is physically meaningful. All in all, endmembers' number estimation and unmixing is yielded jointly by the proposed sparse and low-rank NMF approach. To the best of our knowledge, this is the first work that encapsulates those two problems simultaneously in a single task. The newly formulated minimization problem is efficiently tackled via an alternating proximal Newton-type algorithm. Results obtained on simulated and real data demonstrate the favorable properties of the proposed algorithm.



\section{Problem Formulation}
Let us assume that the pixels of the HSI are mixed according to the linear mixing model (LMM),\begin{align}
 \mathbf{Y} = \boldsymbol{\varPhi}\boldsymbol{\mathcal{W}}^T + \mathbf{E}
 \label{eq:lmm}
\end{align}
where matrix $\mathbf{Y} \in \mathcal{R}_{+}^{L\times K}$ (with $\mathcal{R}^{L\times K}_{+}$ being the $L\times K$ dimensional nonnegative orthant of ${\mathcal R}^{L \times N}$) consists of the $K$ pixels' $L\times 1$ spectral signatures (with $L$ being the number of spectral bands), $\boldsymbol{\varPhi}\in\mathcal{R}_{+}^{L\times N}$ is the endmembers' matrix containing the spectral signatures of the $N$ endmembers and, $\boldsymbol{\mathcal{W}}\in \mathcal{R}_{+}^{K\times N}$ is the matrix containing the abundance fractions of the pixels of the HSI. Finally, $\mathbf{E}\in\mathcal{R}^{L\times K}$ is additive i.i.d Gaussian noise, which contaminates the pixels' spectral information.  

As mentioned earlier, traditional hyperspectral unmixing algorithms assume the number $N$ of endmembers known beforehand. Typically, this number $N$ is provided by an independent procedure which is applied before the unmixing process. In this work, we depart from this rationale, accounting in our methodology also for the lack of knowledge of the actual number of endmembers existing in a hyperspectral scene. More specifically, we overstate the latent dimension $N$ of the matrix product in (\ref{eq:lmm}) and the LMM is subsequently expressed as,
\begin{align}
 \mathbf{Y} = \boldsymbol{\Phi}\mathbf{W}^T + \mathbf{E},
\label{eq:overestimated_LMM}
\end{align}
where $\boldsymbol{\Phi}\in\mathcal{R}_+^{L\times r}$ and $\mathbf{W}\in\mathcal{R}_+^{K\times r}$ are now the endmembers' and abundance matrices respectively, with their relevant dimension overestimated i.e. $r\geq N$.

Non-negative matrix factorization has been widely applied for attacking HU under the LMM, formulating the problem as follows:
\begin{align}
\underset{\boldsymbol{\Phi}\geq 0,\mathbf{W} \geq 0}{\mathrm{min}} \|\mathbf{Y} - \boldsymbol{\Phi}\mathbf{W}^T\|^2_F 
\label{eq:nmf}
\end{align} 
where $\|\cdot\|_F$ denotes the Frobenious norm. Several other constraints may be included in the NMF cost function, giving rise to disparate algorithms. All these constraints aim at capturing inherent structures on the sought matrices such as sparsity, structured sparsity etc, \cite{l21nmf_2011}, stemming from physical properties e.g. spatial correlation. In the case of the rank overestimated LMM of (\ref{eq:overestimated_LMM}), it is obvious that this approach calls for an appropriate regularization term which should be added in (\ref{eq:nmf}). Concretely, such a term should penalize properly the rank of the data representation matrix $\boldsymbol{\Phi}\mathbf{W}^T$. 
 
In light of the above, the ubiquitous tight upper bound of the nuclear norm i.e., 
\begin{align}
 \|\boldsymbol{\Phi}\mathbf{W}^T\|_{\ast} = \mathrm{inf} \frac{1}{2}\left(\|\boldsymbol{\Phi}\|^2_F +  \|\mathbf{W}\|^2_F \right)
\end{align}
has come into play in several works in the literature for imposing low-rankness in bilinear terms of the form  $\boldsymbol{\Phi}\mathbf{W}^T$. The minimization of this upper bound though favors low-rankness by inducing smoothness on matrices $\boldsymbol\Phi$ and $\mathbf{W}$, i.e., it produces matrices with linear dependent columns/rows. Since HU is an instance of blind source separation, the outcome of the above-mentioned bound minimization lies in the opposite direction of what we actually desire. Ideally, we need an algorithm that returns only distinct endmembers' spectral signatures and the respective abundance fractions' of the pixels. 

Considering this and the fact that only a subset of the endmembers in $\boldsymbol{\Phi}$ composes the spectral signature of each pixel (i.e., $\mathbf{W}$ is expected to be sparse), we formulate HU as the following {\it sparse and low-rank} NMF-type optimization problem,
\begin{align}
 \underset{\boldsymbol{\Phi}\geq 0,\mathbf{W}\geq 0}{\mathrm{min}} &\|\mathbf{Y} - \boldsymbol{\Phi}\mathbf{W}^T\|^2_F + \delta \sum^r_{i=1}\sqrt{\|\boldsymbol{\phi}_i\|^2_2 + \|\boldsymbol{\mathit{w}}_i\|^2_2} \nonumber \\
& + \lambda_1 \|\mathbf{W}\|_1.
 \label{cost_function}
\end{align}
The first term in (\ref{cost_function}) performs the fitting between data matrix $\mathbf{Y}$ and its bilinear representation  $\boldsymbol{\Phi}\mathbf{W}^T$, the second term is a novel low-rank promoting penalty and the last term in (\ref{cost_function}) is the $\ell_1$ sparsity inducing norm. Parameters $\delta$ and $\lambda_1$ are the regularization coefficients of the low-rank and sparsity terms respectively. In a few words, (\ref{cost_function}) defines the cost function of a multiple constrained optimization problem, which accounts simultaneously for a) non-negativity of the factors  $\boldsymbol{\Phi}, \mathbf{W}$, b) low-rank on the product $\boldsymbol{\Phi}\mathbf{W}^T$  and c) sparsity on the abundance matrix $\mathbf{W}$. \vspace{0.2cm} \\
\textit{Remark 1: The introduced low-rank promoting term in (\ref{cost_function}) is non-smooth and induces non-separability w.r.t. columns $\boldsymbol{\phi}_i$ and $\boldsymbol{\mathit{w}}_i$ of $\boldsymbol{\Phi}$ and $\mathbf{W}$ respectively. Actually, it is tantamount to applying the column-sparsity promoting $\ell_2/\ell_1$ norm on the augmented  matrix $ [\begin{smallmatrix} \boldsymbol{\Phi} \\ \mathbf{W} \end{smallmatrix} ] $.}

From Remark 1, it can be easily understood that the minimization of the low-rank promoting term of (\ref{cost_function}) results to zeroing jointly columns of $\boldsymbol{\Phi}$ and $\mathbf{W}$. This way, we can claim that the overestimated dimension $r$ decreases and the remaining non-zero columns of the matrices constitute the bases of the subspaces spanned by $\boldsymbol{\Phi}$ and $\mathbf{W}$. 

Next we present a proximal alternating minimization algorithm for solving (\ref{cost_function}).
\section{Proposed Minimization algorithm}
The minimization of (\ref{cost_function}) is by no means a straightforward task. First, exact minimization w.r.t. matrices $\boldsymbol{\Phi}$ and $\mathbf{W}$ is impossible due to the induced non-separability of the low-rank promoting term (Remark 1). Second, joint non-smoothness and non-separability of this term might lead us to irregular points, i.e. to coordinate-wise minima that may not be local minima of the cost function, \cite{Hong_2016}. In an effort to circumvent those two impediments a) we assume a smooth approximation of the low-rank promoting term by adding a small constant $\eta^2$ and b) we replace the smooth terms of (\ref{cost_function}) with quadratic approximate functions, \cite{Hong_2016}. 

Let us first denote the smooth part of (\ref{cost_function}) as,
\small
\begin{align}
 f(\boldsymbol{\Phi},\mathbf{W}) = \frac{1}{2}\|\mathbf{Y} - \boldsymbol{\Phi}\mathbf{W}^T\|^2_F + \delta \sum^r_{i=1} \sqrt{\|\boldsymbol{\phi}_i\|^2_2 + \|\boldsymbol{\mathit{w}}_i\|^2_2 + \eta^2}
\end{align}
\normalsize
and with $g(\mathbf{W}) = \lambda_1\|\mathbf{W}\|_1$ its non-smooth separable part. 
Considering the matrices $\boldsymbol{\Phi}$ and $\mathbf{W}$ as blocks in our problem, each one of them may be updated by alternatingly minimizing the following approximate cost functions,  
\begin{align}
 {\mathbf{W}}^k = & \underset{\mathbf{W} \geq 0}{\mathrm{argmin}} \langle \hat{\mathbf{G}}_{\mathbf{W}}^{k}, \mathbf{W} - \hat{\mathbf{W}}^{k-1} \rangle  \nonumber\\
& + \frac{1}{2\alpha^k}\mathrm{trace}\{(\mathbf{W} - \hat{\mathbf{W}}^{k-1})^T\hat{\mathbf{H}}^{k}_{\mathbf{W}}(\mathbf{W} - \hat{\mathbf{W}}^{k-1})\}\nonumber  \\
& + g(\mathbf{W})
\label{eq:update_W}
\end{align} and 
\begin{align}
 {\boldsymbol{\Phi}}^k = & \underset{\boldsymbol{\Phi} \geq 0}{\mathrm{argmin}} \langle \hat{\mathbf{G}}_{\boldsymbol{\Phi}}^{k}, \boldsymbol{\Phi} - \hat{\boldsymbol{\Phi}}^{k-1} \rangle \nonumber\\ 
    &+ \frac{1}{2\alpha^k}\mathrm{trace}\{(\boldsymbol{\Phi} - \hat{\boldsymbol{\Phi}}^{k-1})^T\hat{\mathbf{H}}^{k}_{\boldsymbol{\Phi}}(\boldsymbol{\Phi} - \hat{\boldsymbol{\Phi}}^{k-1})\}
\label{eq:update_Phi}
\end{align}
where \small$\hat{\mathbf{G}}_{\mathbf{W}}^k =  \nabla_{\mathbf{W}} f(\hat{\boldsymbol{\Phi}}^{k-1}, \hat{\mathbf{W}}^{k-1})$, $\hat{\mathbf{G}}_{\boldsymbol{\Phi}}^k =  \nabla_{\boldsymbol{\Phi}} f(\hat{\boldsymbol{\Phi}}^{k-1}, \hat{\mathbf{W}}^{k})$ \normalsize, \small $\hat{\mathbf{H}}^k_{\mathbf{W}} = \nabla_{\mathbf{W}}^2 f(\hat{\boldsymbol{\Phi}}^{k-1}, \hat{\mathbf{W}}^{k-1})$   \ \normalsize, \small $\hat{\mathbf{H}}^k_{\boldsymbol{\Phi}} = \nabla_{\boldsymbol{\Phi}}^2 f(\hat{\boldsymbol{\Phi}}^{k-1}, \hat{\mathbf{W}}^{k}) $ \normalsize, $\langle \cdot,\cdot\rangle$ denotes inner matrix product and $k$ is the iteration index. Minimization of (\ref{eq:update_W}) as such gives rise to a scaled proximal operator, \cite{prox_newton}, in the form 
\begin{equation}
 \mathbf{W}^{k} = \mathrm{prox}^{{\hat{\mathbf{H}}^{k}}_{\mathbf{W}}}_{\|\cdot\|_1}\left(\mathbf{W}^{k-1} - (\hat{\mathbf{H}}_{\mathbf{W}}^{k})^{-1}\hat{\mathbf{G}}_{\mathbf{W}}^k \right). 
 \label{w_proximal}
\end{equation}
The computation of (\ref{w_proximal}) has given way to disparate approximate or iterative schemes (proximal Newton methods), \cite{prox_newton}, since there exists no closed form solution for non diagonal\footnote{Note that for ${\hat{\mathbf{H}}^{k}}_{\mathbf{W}}$ diagonal, the scaled proximal operator reduces to the known proximity operator of the $\ell_1$ norm i.e. the soft-thresholding operator.} ${\hat{\mathbf{H}}^{k}}_{\mathbf{W}}$ as in our case. Herein, we propose to apply an incremental strategy, \cite{bertsekas2011incremental}, for approximately solving (\ref{eq:update_W}). More specifically, a gradient step is first applied on the smooth part of (\ref{eq:update_W}). The outcome of the gradient step is then provided as input to the proximal operator of the non-smooth term i.e., the $\ell_1$ norm, whose output is finally projected to the feasible set. Regarding the update of $\boldsymbol{\Phi}$, it is a much simpler case due to the absence of non-smooth terms. In general for the parameter $\alpha^k$ in eqs. (\ref{eq:update_W}), (\ref{eq:update_Phi}) it holds $\alpha^k\in(0,1]$ and by setting $\alpha^k=1$ we are led to  pure Newton updates for the smooth parts. Overall, $\boldsymbol{\Phi}$ and $\mathbf{W}$ are computed by using the following expressions,
\small
\begin{align}
\mathbf{W}^k = \mathcal{P}_{\mathcal{R}^{K\times r}_{+}}\left(ST\left(\left(\hat{\boldsymbol{\Phi}}^{T,k-1}\hat{\boldsymbol{\Phi}}^{k-1} + \hat{\mathbf{D}}^{k-1}\right)^{-1}\hat{\boldsymbol{\Phi}}^{k-1}\mathbf{Y},\lambda_1\right)\right)
\label{eq:final_w}
\end{align}
\normalsize
\begin{align}
\boldsymbol{\Phi}^k = \mathcal{P}_{\mathcal{R}^{N \times r}_{+}}\left(\left(\hat{\mathbf{W}}^{T,k}\hat{\mathbf{W}}^{k} + \hat{\mathbf{D}}^{k-1}\right)^{-1}\hat{\mathbf{W}}^{k}\mathbf{Y}\right)
\label{eq:final_phi}
\end{align}
where $ST(x,\lambda) = \mathrm{sign}(x)\mathrm{max}(|x| - \lambda )$ is the soft-thresholding operator and $\mathcal{P}_{\mathcal{R}^{K\times r}_{+}}$ is the projection operator onto $\mathcal{R}_{+}^{K\times r}$. In addition, $\hat{\mathbf{D}}^k$ is a $r\times r$ diagonal matrix with elements 
\begin{align}
\hat{d}^k_{ii} = \frac{\delta}{\sqrt{\|\hat{\boldsymbol{\phi}}^k_i\|^2_2 + \|{\hat{\boldsymbol{\mathit{w}}}}^k_i\|^2_2 + \eta^2}}.
\label{eq:update_d}
\end{align}
Note that the small constant $\eta$ added for smoothing purposes in the introduced low-rank regularization term, averts zeros values on the denominator of $\hat{d}_{ii}^k$.
Moreover, for $\alpha^k\in(0,1)$ quadratic approximate functions are upper-bounds of the original cost function (tight for $\alpha^k=1$). That said, the above-described scheme resembles the block successive upper-bound minimization framework of \cite{Hong_2016}. As stated earlier, and since we are dealing with a constrained minimization problem, the updates for $\boldsymbol{\Phi}$ and $\mathbf{W}$  are projected to the feasible set thus accounting for the nonnegativity constraint. In addition, problem (\ref{eq:update_W}) is solved inexactly by the incremental strategy described earlier. In view of these, for ensuring that the cost function decreases at each step\footnote{Sufficient decrease of the cost function at each step is necessary for the convergence analysis of the algorithm, not given here due to space limitations.}, an extrapolation step is needed and the final estimates of $\boldsymbol{\Phi}$ and $\mathbf{W}$ at the $k$th iteration are obtained as 
\begin{align}
\hat{\mathbf{W}}^k = \hat{\mathbf{W}}^{k-1} + \beta_{\mathbf{W}}^k \left(\mathbf{W}^k - \hat{\mathbf{W}}^{k-1}  \right), \\
\hat{\boldsymbol{\Phi}}^k = \hat{\boldsymbol{\Phi}}^{k-1} + \beta_{\boldsymbol{\Phi}}^k \left(\boldsymbol{\Phi}^k - \hat{\boldsymbol{\Phi}}^{k-1}\right).
\end{align}
Typically,  $\beta_{\mathbf{W}}^k$ and $\beta_{\boldsymbol{\Phi}}^k$ are adjusted dynamically so that the cost function's sufficient decrease is guaranteed at each step. Along this line, numerous schemes, known as line search methods have come into play e.g. backtracking. Those schemes affect the convergence and rate of convergence of the algorithms to stationary points. The resulting algorithm is given in Algorithm 1.
\begin{algorithm}
\caption{The proposed sparse and low-rank NMF endmembers' number estimation and HU algorithm}
\footnotesize
 \begin{algorithmic}
 \footnotesize{
  \STATE \bf{Initialize} $\hat{\mathbf{W}}^0,\hat{\boldsymbol{\Phi}}^0,\hat{\mathbf{D}}^0,\beta_{\boldsymbol{\Phi}}^{0},\beta_{\mathbf{W}}^{0}$
	\STATE \bf{Select} $\delta,\lambda_1$
	\STATE \bf{for} $k=1,2,\dots,\mathrm{Max\_iter}$
	 \STATE  $\mathbf{W}^k = \mathcal{P}_{\mathcal{R}^{K\times r}_{+}}\left(ST\left(\left(\hat{\boldsymbol{\Phi}}^{T,k-1}\hat{\boldsymbol{\Phi}}^{k-1} + \hat{\mathbf{D}}^{k-1}\right)^{-1}\hat{\boldsymbol{\Phi}}^{k-1}\mathbf{Y}\right),\lambda_1\right)$
	 \STATE  $\hat{\mathbf{W}}^k = \hat{\mathbf{W}}^{k-1} + \beta_{\mathbf{W}}1^{k-1} \left(\mathbf{W}^k - \hat{\mathbf{W}}^{k-1}  \right)$
	 \STATE  $\boldsymbol{\Phi}^k = \mathcal{P}_{\mathcal{R}^{L \times r}_{+}}\left(\left(\hat{\mathbf{W}}^{T,k}\hat{\mathbf{W}}^{k} + \hat{\mathbf{D}}^{k-1}\right)^{-1}\hat{\mathbf{W}}^{k}\mathbf{Y}\right)$
	 \STATE  $\hat{\boldsymbol{\Phi}}^k = \hat{\boldsymbol{\Phi}}^{k-1} + \beta_{\boldsymbol{\Phi}}^{k-1}  \left(\boldsymbol{\Phi}^k - \hat{\boldsymbol{\Phi}}^{k-1}\right)$
	 \STATE $\hat{d}^k_{ii} = \frac{\delta}{\sqrt{\|\hat{\boldsymbol{\phi}}^k_i\|^2_2 + \|\boldsymbol{\hat{\mathit{w}}}^k_i\|^2_2 + \eta^2}}$, $i=1,2,\dots,r$, $\hat{\mathbf{D}}^k =\mathrm{diag}(\hat{\mathbf{d}}^k)$
	  \STATE  $\mathrm{Update} \ \beta_{\mathbf{W}}^k, \beta_{\boldsymbol{\Phi}}^k$
	   \STATE \bf{end}	 }
 \end{algorithmic}
\end{algorithm}

All in all, the proposed algorithm requires an overestimate of the actual number of endmembers, which relaxes to a large degree the need for knowing in advance the exact number thereof. Then, as the algorithm proceeds, columns of the endmembers' and abundance matrices $\boldsymbol{\Phi}$ and $\mathbf{W}$ are jointly zeroed as a result of the $\ell_2/\ell_1$ norm based low-rank promoting term adopted. The remaining non-zero columns minimize the reconstruction error while at the same time sparsity is induced on the abundance matrix. 
Next we experimentally validate the favorable characteristics of the proposed algorithm i.e., its ability to uncover the true number of the endmembers existing in a hyperspectral scene, along with performing highly accurate blind unmixing.
\begin{figure}[t!]
\centering
 \begin{tabular}{c c}
 \includegraphics[width=0.22\textwidth,height=0.2\textwidth]{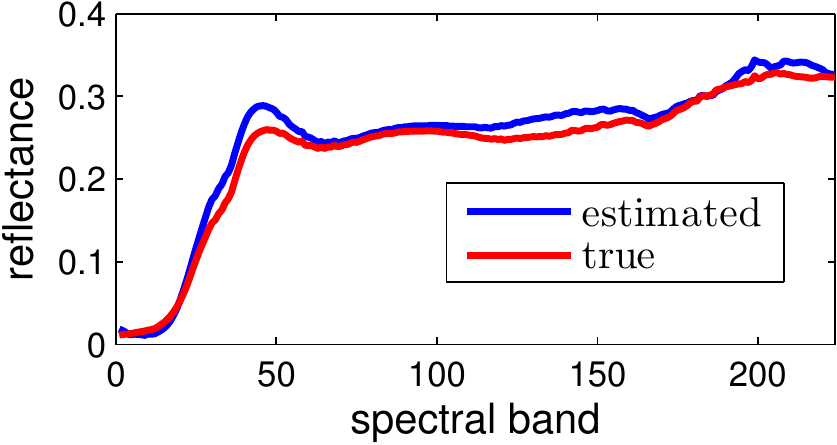}&\includegraphics[width=0.22\textwidth,height=0.2\textwidth]{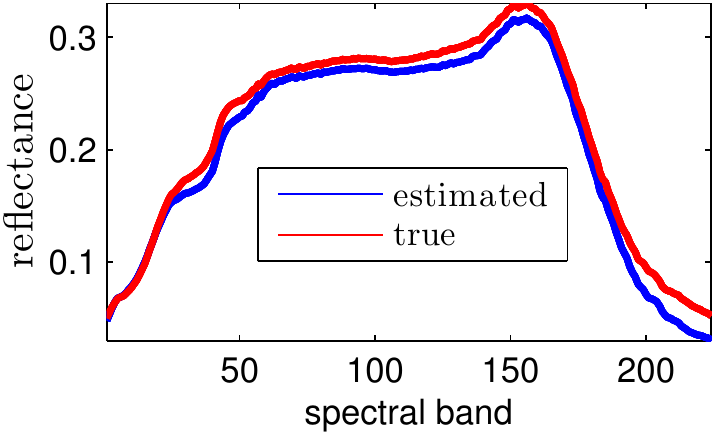} \\
 (a) & (b)  \\
 \includegraphics[width=0.22\textwidth,height=0.2\textwidth]{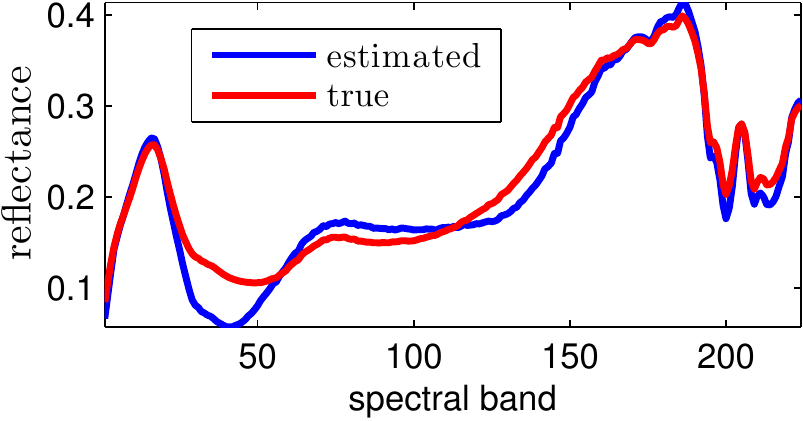} & \includegraphics[width=0.22\textwidth,height=0.2\textwidth]{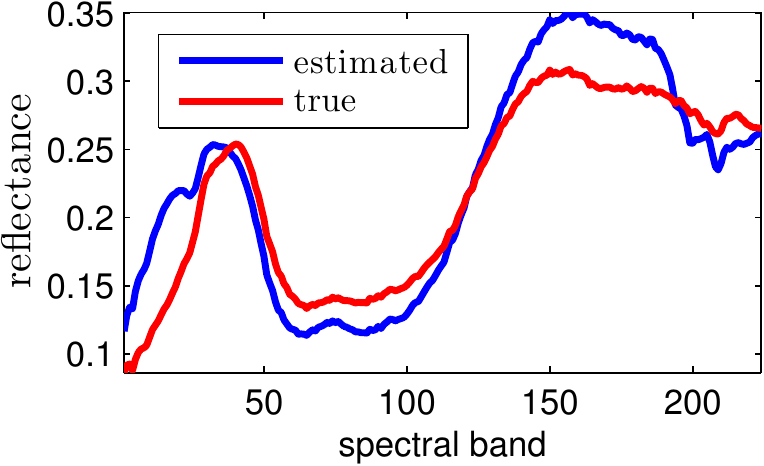} \\
 (c) & (d) 
 \end{tabular}
 \caption{Endmembers' spectral signatures obtained by the proposed algorithm on the simulated data experiment.}
 \label{fig:sim_data}
\end{figure}
\vspace{-0.3cm}
\section{Experiments}
In this section we test the performance of the proposed algorithm to a simulated and a real data experiment.
\subsection{Simulated data experiment}
In this experiment we aim at corroborating the competence of the proposed algorithm in uncovering the true number of endmembers along with estimating the spectral signatures of the endmembers. To this end, we generate a $500\times 4$ abundance  matrix whose elements follow a uniform distribution in the interval [0,1]. This matrix is then sparsified by randomly keeping only 30\% of its elements. From the USGS spectral library, we select randomly 4 endmembers' spectral signatures measured at $L=224$ distinct spectral bands. Then, we linearly produce $K=500$ simulated pixels' spectral signatures under the LMM framework. The pixel spectral signatures are then contaminated by additive i.i.d. Gaussian noise with standard deviation $\sigma=10^{-3}$. 

Since the main premise of our approach is the development of a blind unmixing method that exhibits robustness in the absence of knowledge of the true number of endmembers, we initialize the proposed algorithm with an overestimate $r=10$ of the actual number of endmembers. Both endmembers' and abundance matrices are randomly initialized according to the uniform distribution. Interestingly, the proposed algorithm converges to abundance and endmembers' matrices consisting of 4 non-zero columns, which is the same as the actual number of endmembers that produced the data. Moreover, it can be easily observed from Fig. \ref{fig:sim_data}, that the estimated endmembers' spectral signatures present high degree of similarity to the real ones. Hence, we can conclude that the proposed algorithm is, in principle, capable of carrying out the challenging task of simultaneously estimating the number of endmembers and performing blind unmixing.
\begin{figure}
\begin{tabular}{c c c}
 \includegraphics[width=0.15\textwidth,height=0.4\textwidth]{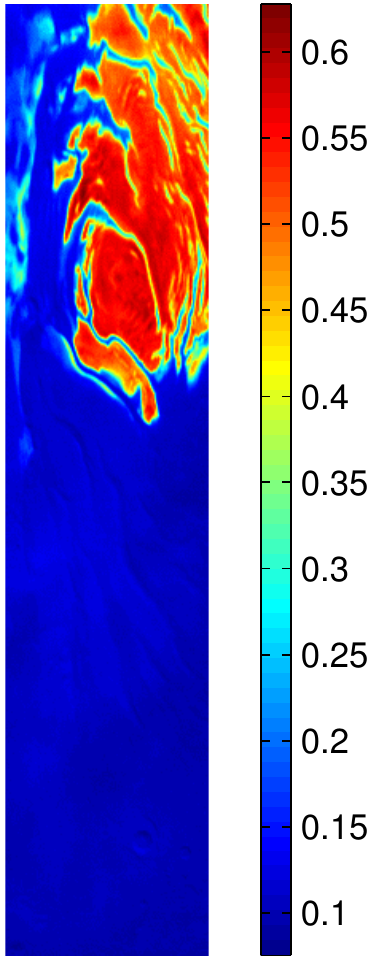} & \includegraphics[width=0.15\textwidth,height=0.4\textwidth]{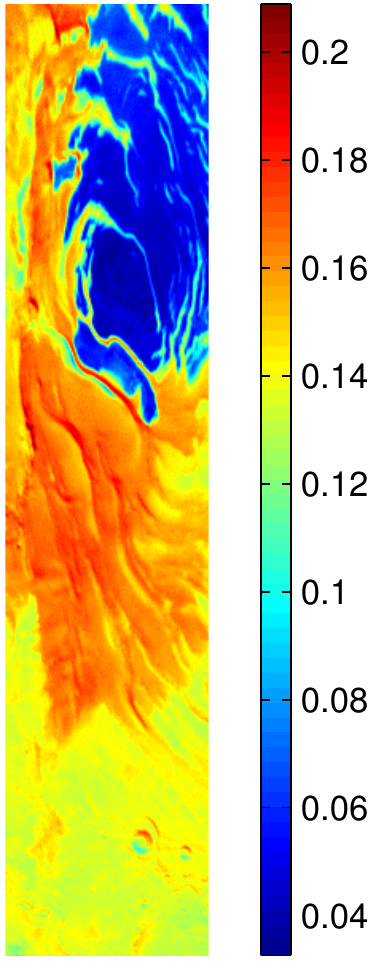} & \includegraphics[width=0.15\textwidth,height=0.4\textwidth]{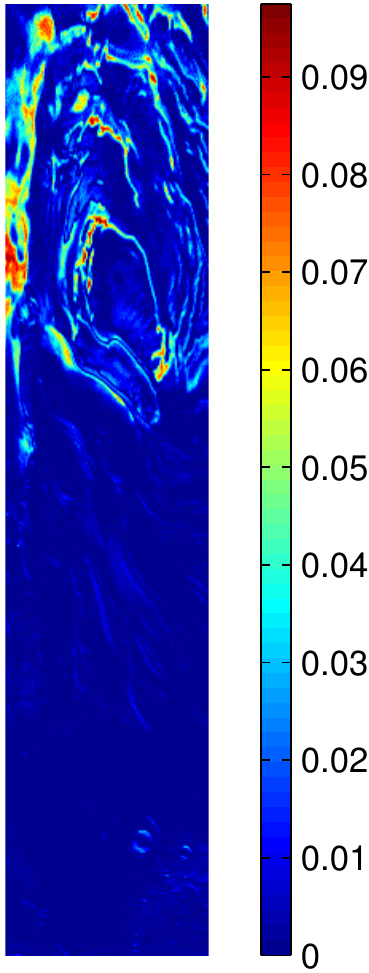} 
\end{tabular}
\caption{Abundance maps of South Polar Cap obtained by the proposed algorithm.}
\label{fig:abund_maps}
\end{figure}
\begin{figure}
 \begin{tabular}{c c}
\includegraphics[width=0.23\textwidth,height=0.2\textwidth]{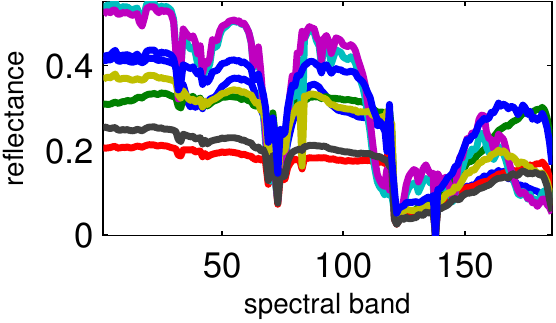} &  \includegraphics[width=0.23\textwidth,height=0.2\textwidth]{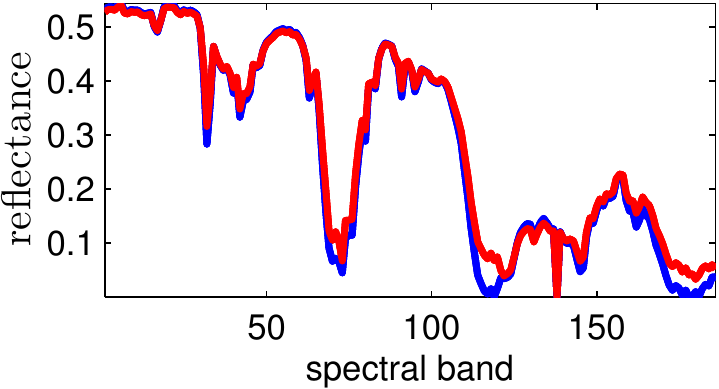}\\
(a)  & (b) $CO_2$\\
 \includegraphics[width=0.23\textwidth,height=0.2\textwidth]{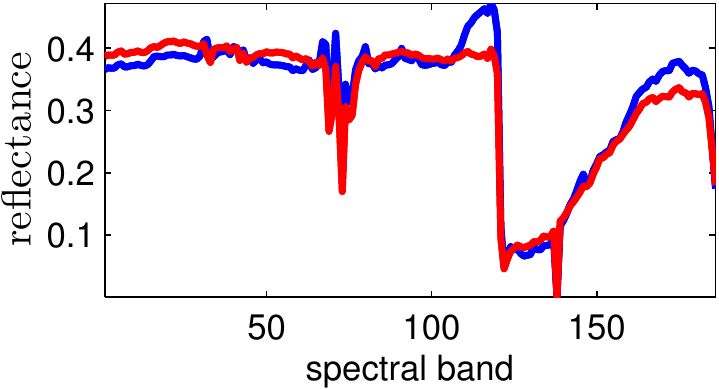} & \includegraphics[width=0.23\textwidth,height=0.2\textwidth]{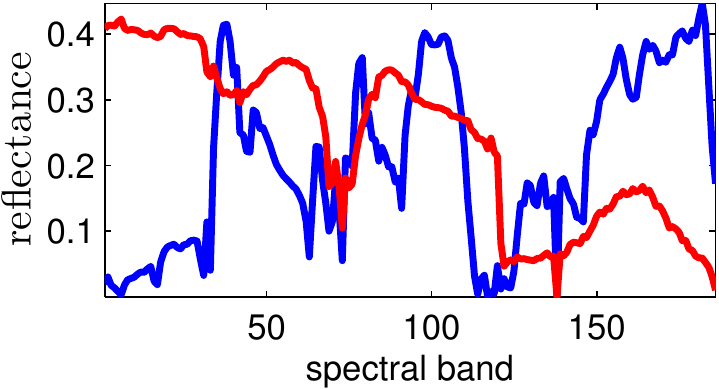}\\
 (c) dust & (d) $H_2O$
 \end{tabular}
 \caption{(a) Eight endmembers' signatures obtained by VCA on South Polar Cap, (b)-(d) endmebers' signatures  estimated by the proposed algorithm (blue lines) and VCA (red lines).}
  \label{fig:vca_proposed}
 \end{figure}
\vspace{-0.0cm}
\subsection{Real data Experiment}
Herein our goal is to test and validate the proposed algorithm on a real hyperspectral dataset. To this end we utilized the OMEGA ORB0041 image (known as South Polar Cap) which covers a large part of the south polar of Mars. Previous studies, e.g. \cite{themelis2012unmixing}, reported three principal chemical species on the surface: $H_2O$ ice, $CO_2$ ice, and mineral dust. 

To evaluate the performance of the proposed algorithm on this real hyperspectral dataset, we suitably initialized the endmembers' matrix with the outcome of the VCA algorithm, \cite{nascimento2005vertex}, which was run with an overestimate $r=8$ of endmembers. As is shown in Fig. \ref{fig:vca_proposed}(a), the resulting by VCA endmembers' matrix contains correlated spectra, which is expected from our prior knowledge i.e, that there exist 3 (instead of 8) endmembers in this specific image. Since there exist no reference spectra of the actual endmembers' spectral signatures, for comparison purposes, we use the ones returned by VCA which, this time, was run for the correct number of endmembers.  Notably, the proposed algorithm is proven to be capable of estimating the actual number of the endmembers.
Moreover, the estimated endmembers' signatures (Fig. 3(b)-3(d)) for $CO_2$ and dust are quite close to those resulting by VCA, while the opposite holds for the respective spectral signature of the $H_2O$. This is probably due to the small abundance values of $H_2O$, as shown in Fig. 2(c). Additionally, as it can be observed in Fig. \ref{fig:abund_maps}, the resulting abundance maps are quite close to those that have been published in literature, \cite{themelis2012unmixing}. 
\vspace{-0.3cm}
\section{Conclusions}
In this paper, the problems of a) estimating the number of endmembers and b) blind unmixing are formulated as a joint optimization task. The ambitious procedure is carried out by first formulating the problem via a novel $\ell_2/\ell_1$ norm based low-rank promoting term. Then, an alternating proximal Newton based algorithm is proposed for minimizing the emerging cost function. Promising results obtained on simulated and real data experiments corroborate the merits of this novel approach. 

%
%
\bibliographystyle{IEEEtran}
\bibliography{IEEEabrv,refs_report}

%




\end{document}